\documentclass[10pt,twocolumn,letterpaper]{article}

\usepackage{cvpr}
\usepackage{times}
\usepackage{graphicx}
\usepackage{amsmath}
\usepackage{amssymb}

\usepackage{siunitx}
\usepackage{float}
\usepackage{stfloats}
\usepackage{color}
\usepackage{lipsum}
\usepackage[export]{adjustbox}
\usepackage[backend=biber,sorting=ynt,maxnames=1]{biblatex}
\usepackage[utf8]{inputenc}
\usepackage{booktabs}

\usepackage[autostyle, english = american]{csquotes}
\MakeOuterQuote{"}

\newcommand{\fakesection}[1]{%
  \par\refstepcounter{section}
  \sectionmark{#1}
  \addcontentsline{toc}{section}{\protect\numberline{\thesection}#1}%
}
\newcommand{\review}[1][]{\textsuperscript{\color{red} [#1]}}
\newcommand{\citneeded}[1][]{\review[citation needed]}
\newcommand{\citay}[1][]{\citeauthor{#1} (\citeyear{#1})}

\DefineBibliographyStrings{english}{%
    andothers = {\em et\addabbrvspace al\adddot}
}
\DeclareCiteCommand{\citeauthor}
  {\boolfalse{citetracker}%
   \boolfalse{pagetracker}%
   \usebibmacro{prenote}}
  {\ifciteindex
     {\indexnames{labelname}}
     {}%
   \printtext[bibhyperref]{\printnames{labelname}}}
  {\multicitedelim}
  {\usebibmacro{postnote}}

\bibliography{references}

\usepackage[breaklinks=true,bookmarks=false]{hyperref}

\cvprfinalcopy

\def\cvprPaperID{****} 

\begin{document}


\title{Implementing Adaptive Separable Convolution\\
for Video Frame Interpolation}


\author{%
Mart Kartašev\thanks{Equal contributions}\qquad%
Carlo Rapisarda\footnotemark[1]\qquad%
Dominik Fay\footnotemark[1]\\
KTH Royal Institute of Technology\\
{\tt\small \{kartasev, carlora, dominikf\} @kth.se}}

\maketitle


\begin{abstract}
\label{sec:abstract}

As Deep Neural Networks are becoming more popular, much of the attention is being devoted to Computer Vision problems that used to be solved with more traditional approaches. Video frame interpolation is one of such challenges that has seen new research involving various techniques in deep learning. In this paper, we replicate the work of \textnormal{\citeauthor{SepConv}} on Adaptive Separable Convolution, which claims high quality results on the video frame interpolation task. We apply the same network structure trained on a smaller dataset and experiment with various different loss functions, in order to determine the optimal approach in data-scarce scenarios. The best resulting model is still able to provide visually pleasing videos, although achieving lower evaluation scores.
\end{abstract}


\section{Introduction}
\label{sec:introduction}

Video Frame interpolation, also known as motion-compensated frame interpolation (MCFI) - motion interpolation for short - has long been an area of research in the field of Computer Vision. In simple terms, it is the process of generating an intermediate image between two frames of a video by processing them with an interpolation technique. The primary motivation for this is usually to make the video seem more smooth or fluid, as well as reducing the effects of motion blur. The main goal then is to generate high quality frames in order to increase the frame-rate of the video, without creating visible distortions. There are other areas of application, such as matching the frame-rate of a video to that of display hardware. Such techniques are sometimes run offline (i.e. the frames are pre-processed), but also occasionally on-the-fly, as is the case in some modern displays, creating a demand for computationally efficient solutions.

Traditionally, frame interpolation is solved by first modelling motion between images explicitly, and then synthesizing the intermediate image from the motion representation. Most commonly, optical flow is used for the former. Optical flow describes, for every pixel, the direction (2D) and magnitude of movement between the two images.
In the context of neural networks, and especially deep learning, the problem is often solved with the help of convolutional neural networks. 

The subject network described in this paper has the goal of generating intermediate frames based on the immediate successor and predecessor frames.
It is important to note that this work is in essence a replication of the work of \citeauthor{SepConv}. From the start, we had the plan of attempting to use the same structure on a downscaled dataset and investigating the effect of different loss functions on the training.
The network we create is intended to be an end-to-end solution which can take any given video as an input and generate a sequence of frames doubling the original frame-rate.


\section{Background}
\label{sec:related_work}
As mentioned earlier, traditional methods use optical flow for interpolation. However, in general, optical flow cannot be calculated from images without ambiguity (known as the \textit{aperture problem} in computer vision); additional constraints are needed to find a unique solution. Therefore, the quality of the interpolation heavily depends on the accuracy of the flow estimation. Examples of such additional assumptions include brightness constancy and phase constancy. Unfortunately, these assumptions are often violated by scenes with difficult lighting conditions.

Early attempts to alleviate these problems introduce heuristics and/or regularization, which, in turn, impose other limitations, such as the amount of motion that can be handled. Alternatively, methods that entirely operate in the phase domain have been used. A recent successful method \cite{PhaseBased} uses multi-layered pyramids to represent phase information at different resolution levels. It has been found to work well in most cases but lacks robustness in situations with large changes between the frames.

More recently, convolutional neural networks have been used in an attempt to increase robustness.
Since the optical flow ground truth is generally unavailable, supervised learning is usually performed on the frames directly. This comes with the additional benefit of end-to-end trainability. \citay[ImageMatching] perform frame interpolation as an intermediate step for image matching. However, they report that their predictions are often blurry \cite{ImageMatching}. \citay[AdapConv] approach frame interpolation as the task of predicting a local convolution kernel for each pixel in the two frames. In their first work, they use a fully convolutional network to predict 2D convolution kernels \cite{AdapConv}. In a second work, they improve their results by approximating the 2D kernels by two 1D kernels which, as they report, greatly improves memory and computational efficiency. This allows them to use a higher kernel size, thus increasing the amount of motion that can be captured. Furthermore, up- and downsampling layers as well as skip connections are added to the architecture, thus forming a U-Net \cite{SepConv}. This work is the main subject of this paper and is presented in detail in section \ref{sec:approach}.

There are several other CNN-based approaches published at the same or a later time that have also produced strong results. \citay[DeepVoxelFlow2017] propose to perform trilinear interpolation between the two frames, based on what they term voxel flow, which describes movement on the voxel level\footnote{Here, the term voxel refers to one of the color values of a pixel. Typically, a pixel consists of three voxels (R, G and B).}, rather than the pixel level. A CNN is used to predict the voxel flow. \citay[SuperSloMo] focus on producing very high-framerate videos by predicting many intermediate frames at once. They suggest two separate U-Nets to predict and refine both bi-directional optical flow and a visibility map, both of which are used to synthesize intermediate frames \cite{SuperSloMo}. \citay[MultiScaleDeepLossAndGAN] use a multi-scale GAN for interpolation. Every scale captures a different level of detail and all scales are trained jointly with a loss function that incorporates both a content loss and an adversarial loss \cite{MultiScaleDeepLossAndGAN}. A context-aware approach was proposed by \citay[ContextAware2018]. Their suggestion is to incorporate per-pixel contextual information in the image synthesis process. The contextual information is provided by a pre-trained ResNet-18 network \cite{ResNet2016}. Finally, \citay[PhaseNet] use a decoder network that learns to estimate the phase decomposition of the intermediate frame in the form of a steerable pyramid \cite{SteerablePyramid1995}. Subsequently, the phase decomposition is reversed to restore the intermediate image \cite{PhaseNet}.

Methods similar to the above have been used in closely related tasks. In fact, frame interpolation is just a special case of image-based rendering, where middle frames are interpolated from temporally neighboring frames. Other forms of image-based rendering include view synthesis, style transfer and frame extrapolation.

\begin{figure*}
    \begin{center}
    	\includegraphics[width=\textwidth]{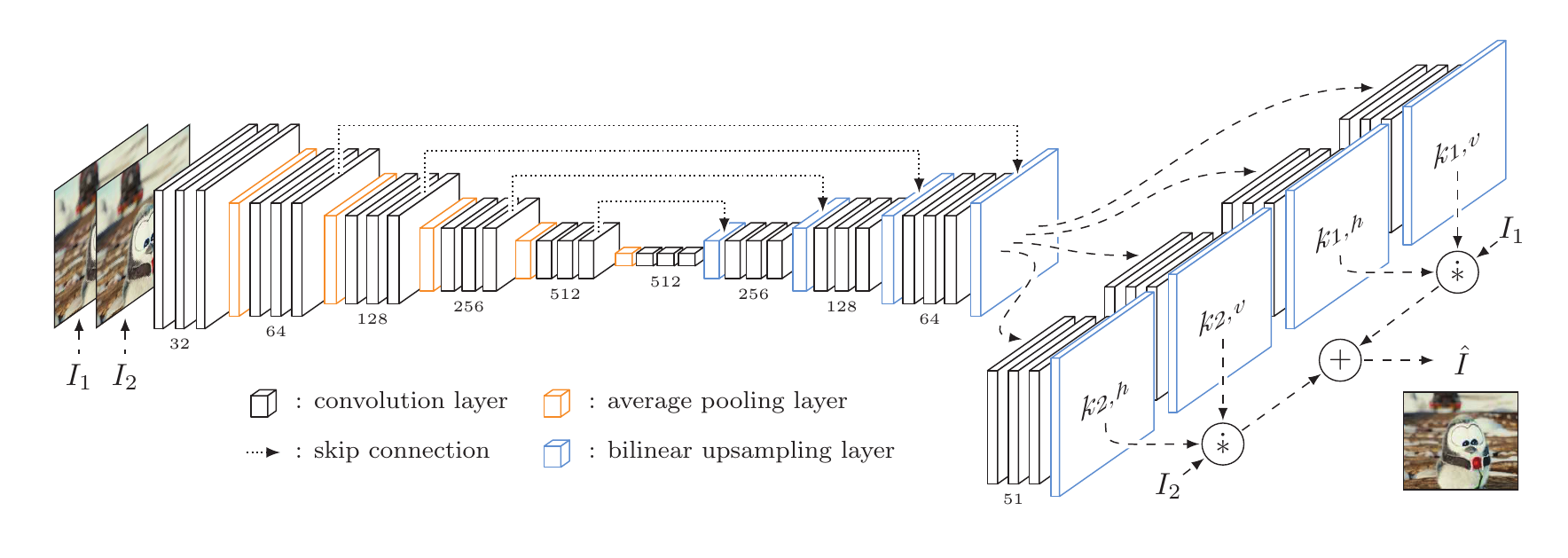}
    \end{center}
    \caption{Structure of the implemented network. Image from \citeauthor{SepConv}}
    \label{fig:networkstructure}
\end{figure*}


\section{Approach}
\label{sec:approach}


\subsection{Network structure}
The structure of our implementation is identical to that of Adaptive Separable Convolution \cite{SepConv}, depicted in figure \ref{fig:networkstructure}. As evidenced by the image, the network has a u-shape structure, composed of several modules with skip connections. Each module contains a pooling or upsampling layer as well as three convolutional layers.

The input to the network is two $128$x$128$ pixel image patches which are cropped from random locations within larger frames. This means that one data point consists of a triplet - the ground truth image as well as its predecessor and successor frames. Both images are given in the RGB colorspace and combined into a tensor with 6 channels. The values available in figure \ref{fig:networkstructure} below each module show how the number of channels in the network changes with each pooling and upsampling layer. All the convolutional layers in the modules use a stride of 2 and kernel size of 2x2 and are activated using ReLu.

The unique property of this network is found in the final part. The network branches into four sub-networks, each of which predict a 1D local convolution kernel to be convolved with the input frames. \textit{Local} in this context refers to the fact that a different kernel matrix is used for every pixel of the input frames. The predicted kernel for a pixel therefore provides the coefficient for a weighted sum of neighboring pixels in the input frames and is a representation of the local motion between the frames. Furthermore, each pair of 1D kernels is used to approximate a 2D kernel as one kernel is applied horizontally and the other vertically. This results in a large saving of memory and computational power without constraining the amount of that kernels can be represented. The predicted middle frame is then the sum of two local convolutions.

It is important to stress that the aforementioned kernels are not weights of the network, but its output. That is, different kernels will be obtained for every pair of input images. Note that this makes intuitive sense because the kernels are per-pixel representations of the motion between the frames, so they should be dependent on the input images.


\subsection{Loss functions}
During the course of this project we experimented with multiple different loss functions for our network. The primary loss function used for training was the sum of absolute differences, also known as $L_1$. Many papers also reference the mean squares loss, known as $L_2$, but cite it as being generally less reliable in the context of frame interpolation -- resulting in blurrier images. We calculate the $L_1$ loss as the absolute difference between the values of the individual pixels as described in eq. \ref{eq:l1}.

\begin{equation}
L_1= \|I - I_{gt}\|_1
\label{eq:l1}
\end{equation}

In addition to the $L_1$ loss, we also tried to use the VGG-19 \cite{VGG} network as a feature extractor, and a loss based on SSIM for fine-tuning. SSIM is normally used as a similarity measure, but it can also be used as a loss function because it is differentiable. We elaborate further on the use of these loss functions in section \ref{sec:training}, and focus on their definitions here.

The use of VGG is motivated by the need to recognize higher level image features that we wish to interpolate in the image, as opposed to simply comparing the differences between frames pixel-by-pixel. This can help create visually pleasing and sharper images, but needs to be used with caution, as it can overtake the $L_1$ loss which ensures a better overall interpolation in terms of pixel positioning.

We use the output from the ReLu activation layer of the convolutional layer in the 4th module of VGG-19 as a feature extractor. This is used in a weighted form together with the $L_1$ loss, to form what we call combined loss for fine tuning. In our early experiments with VGG, we saw that it was approximately a factor of $10^5$ larger than the $L_1$, thus combining it required a weighing factor in order for VGG not to completely overtake the $L_1$ loss. The combined loss was formulated as in eq. \ref{eq:combined}.

\begin{equation}
L_{cmb}= L_1 + \nu \cdot VGG_{19}
\label{eq:combined}
\end{equation}

Additionally, our experiments featured SSIM as a loss function. SSIM is a differentiable similarity measure that is normally used to evaluate generated images for various purposes. We used a window size of 11 in order to calculate the SSIM measure between ground truth and interpolated images.

SSIM and the combined loss are our additions; the other losses have also been tested by \citeauthor{SepConv}.


\subsection{Implementation}
The project is implemented in Python, specifically using PyTorch, as it is widely used in the field, and has a rapidly growing community around it; this was also the framework of choice for \citeauthor{SepConv}. Most layers are standard convolutions, pooling layers, or upsampling operations, therefore it's possible to use the built-in modules of PyTorch. On the other hand, the last operation -- the local convolution between the inputs and the separable filters -- requires a custom extension, and was implemented in CUDA by \citeauthor{SepConv}. In addition, we utilize a fork of the original repository made available by \citeauthor{GibbonsFork}, which extends the CUDA module further by implementing the backward pass, rendering the network training tractable. This was necessary because the code for the gradient calculation had not been published as part of the original implementation. We ensured the correctness of the gradient with the \texttt{gradcheck} function of PyTorch. For debugging and interpolating on CPU, we developed a less optimized version of this module directly in Python.


\subsection{Data}
\label{sec:data}
Similarly to \citeauthor{PhaseNet}, we used triples from the DAVIS dataset \cite{Davis16, Davis17} as our training data. In particular, we run a pre-processor that extracts 150x150 patches from the images, randomly selecting up to 20 from each triplet of frames. Following \cite{SepConv}, we select each patch with a probability that depends on the optical flow between the first and the third frame; specifically, we used the SimpleFlow algorithm implemented in the OpenCV library \cite{OpenCV}. Also, jump-cuts are avoided as suggested in \cite{SepConv}; the researchers do not specify which method is used for this purpose, but for our implementation we used a method based on the histogram differences of the RGB channels, inspired by \citeauthor{Priya10}. With these constraints in place, a total of 16,768 triples were selected for training, all of which were cached on disk to reduce the CPU load when fetching the data.

Even though \citeauthor{PhaseNet} report that about 10k triples was enough to achieve competitive results with their network, \citeauthor{SepConv} state that their architecture was trained with 250k examples. As our network reproduces the latter work, we have implemented several methods to augment our dataset on the fly. In particular, in addition to performing vertical and horizontal flips as in \cite{SepConv}, we also implemented a rotation of $\pm\ang{90}$. Note that these three transforms are applied with equal probability each time a triplet is read, together with a forth option that consists in using the original sample as it is. Independently of the transforms, we also perform the temporal order swap of the first and the third patches of the triplet with a probability of $1/2$, like \citeauthor{SepConv}.


\subsection{Training}
\label{sec:training}
In order to train the network quickly and efficiently, we used a virtual machine hosted on Google Cloud Platform. The hardware configuration used was composed by a Nvidia K-80 GPU (with 12 GB of VRAM), 4 CPU cores, 16 GB of system memory on an Ubuntu 17 OS. Training (with $L_1$) for one epoch took about 19 minutes on DAVIS17 and 12 minutes on DAVIS16. The total training time with DAVIS17 was 16 hours for 50 epochs and with DAVIS16 was 10 hours for 50 epochs.

As optimizer we use AdaMax, with a batch size of 16 and a learning rate of 0.01. The selection of batch size and learning rate was again informed by \citeauthor{SepConv} and served us well in our training. All of our training runs use the network training and validation loss as performance metrics. Additionally, a small selection of 10 images was interpolated with the resulting model after each epoch of training for visual evaluation.


\subsection{Training with DAVIS 2016}

\begin{figure}[t]
	\begin{center}
	\includegraphics[width=\linewidth]{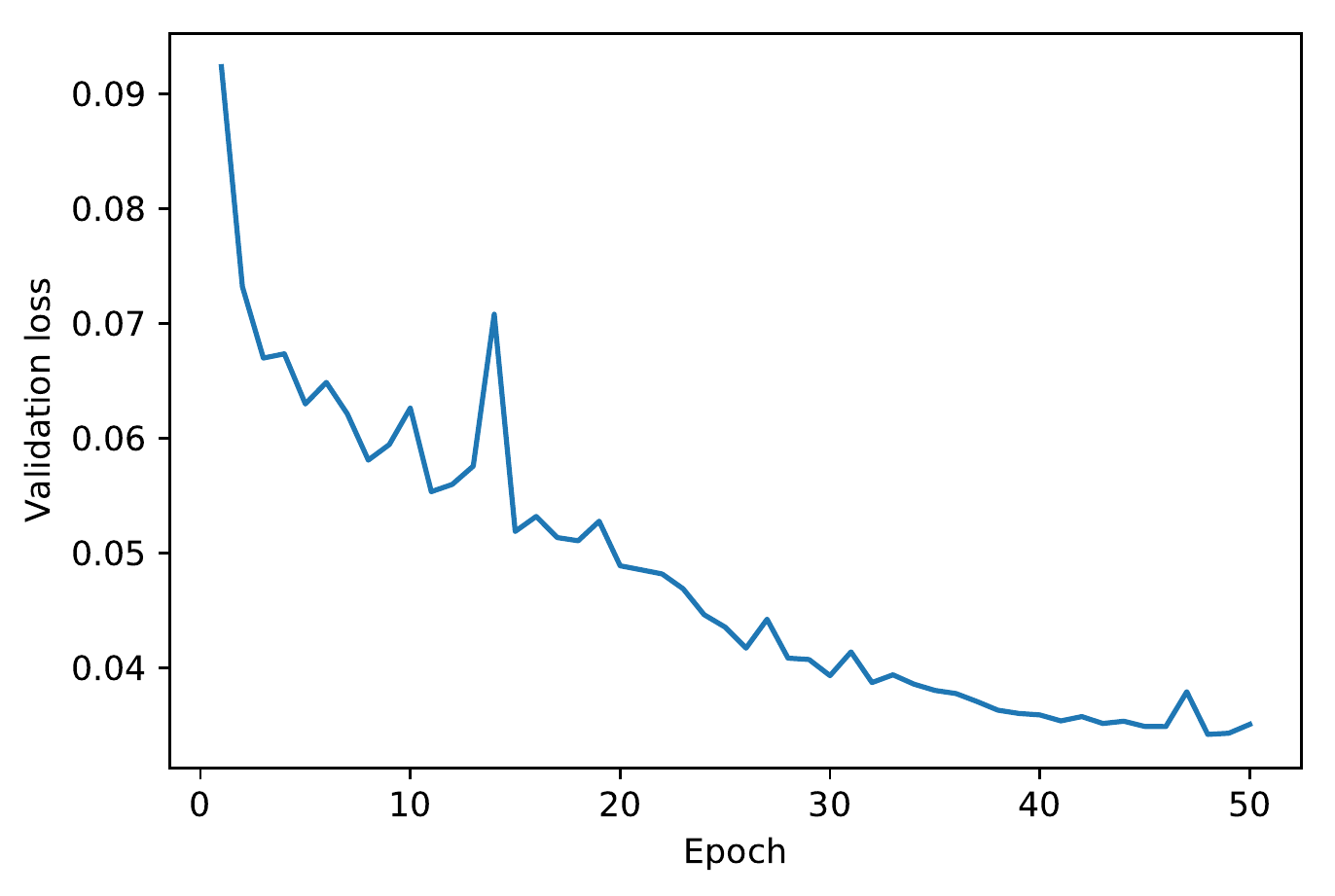}
	\end{center}
	\caption{Plot of the $L_1$ loss for 50 epochs, DAVIS 16}
    \label{fig:davis16training}
\end{figure}

As mentioned in section \ref{sec:data}, we use the DAVIS datasets for training, initially using only the 2016 version -- this first run contained 9952 patches as triplets of frames. The plot of the $L_1$ validation loss used during this training can be seen in image \ref{fig:davis16training}. Swapping of the temporal order was not enabled for this run. 

The final result of this run after a total 50 epochs yielded sub-par results, with the training already plateauing. The interpolation was fundamentally functional, but blurry and generalized poorly.


\subsection{Training with DAVIS 2017}
As we believed that the network was starting to overfit too quickly on the amount of data we had extracted from DAVIS 2016, we increased the amount of examples by using the same pre-processing heuristics on DAVIS 2017, which is a superset of the 2016 version. We used the same parameters as before, but our number of training samples was increased as described in section \ref{sec:data}.

In order to conserve time and resources, we started this training session from the 37th epoch of the previous one. We used an earlier epoch than 50, as we believed that the model had already reached a plateau and had started to overfit the data. This session was also trained with the $L_1$ loss. 


\subsection{Fine-tuning with combined loss}
Once we had reasonably good results from the previous session of training on DAVIS 2017, we decided to try to improve the image quality by implementing different loss functions. For this purpose, we implemented additional metrics, which will be discussed under the relevant sections. We experimented by training the following combinations of losses starting from the 50th epoch of the DAVIS 2017 training session:

\begin{itemize}
    \item $L_1$
    \item $SSIM$
    \item $VGG_{19}$
    \item $L_1$ + $\nu \cdot VGG_{19}$
\end{itemize}

where $\nu$ is a weight to balance the influence between the two components. The results of this training will be further discussed in section \ref{sec:experiments}.

\begin{figure*}
    \begin{center}
    	\includegraphics[width=0.45\textwidth]{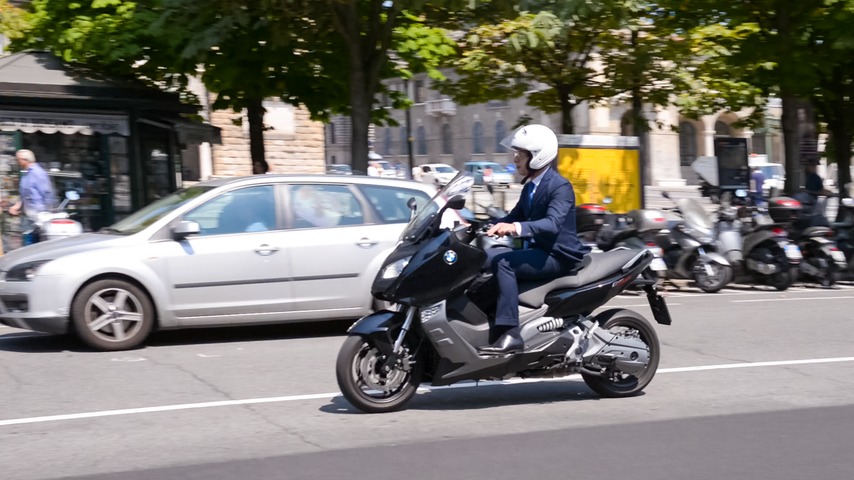}
    	\includegraphics[width=0.45\textwidth]{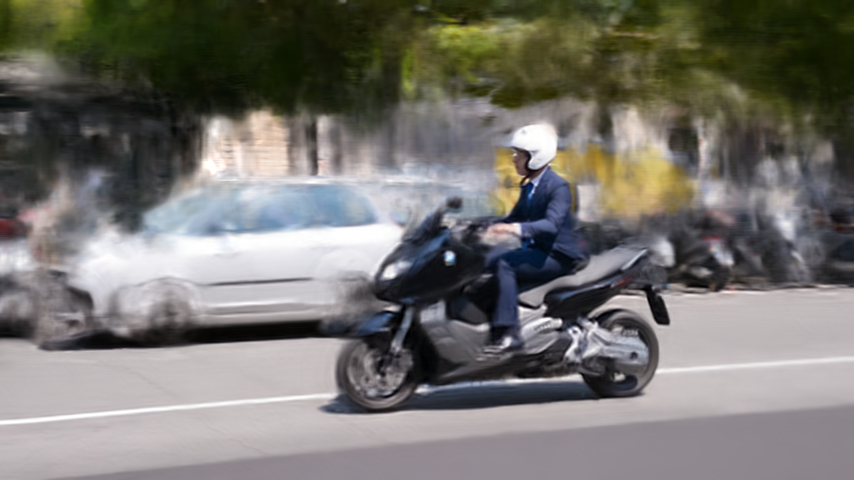}
    \end{center}
    \caption{Ground truth (left) and prediction (right), from "Black scooter" in DAVIS 17, interpolated using the combined loss model.}
    \label{fig:scooter}
\end{figure*}


\section{Experiments}
\label{sec:experiments}

There were two primary goals when evaluating the performance of our trained network. Firstly, we wanted to evaluate the performance of different loss functions in comparison to each other in order to select the best model out of those we trained. Secondly, we wanted to compare the results of our best models against the results of the implementation of \citeauthor{SepConv}.

As the various loss functions do not have comparable ranges, the use of different metrics is necessary to evaluate the models. In order to compare the solutions, we decided to use two common image quality measures, PSNR and SSIM. An important note about these metrics is that they are calculated based on the generated images in comparison with the ground truth.

All the above results were used in addition to visual experiments with each of the involved models. Visual inspection cannot be used directly as a metric, as it is subjective, but it can reveal artifacts and other distortions which are hard to measure with quantitative methods.


\subsection{Results}
\label{sec:results}

The comparison of SSIM and PSNR for the best resulting models from each training session can be seen in table \ref{table:comparisonAllModels}. We can see here that the best objective results were achieved with pure $L_1$ loss. Based on qualitative (visual) inspection, we also picked the combined loss of $L_1$ + VGG with a factor of 1e-5. These are the models we chose to use for the final benchmark runs. Both of them yield better quantitative results than simple linear interpolation. Examples of images used for visual comparison are available in the appendix (\ref{sec:appendix}). 

\begin{table}[h]
\centering
\begin{tabular}{|l|c|c|}
\hline
	Model & SSIM & PSNR \\
	\hline
$L_1$                                       & 0.820 & 29.1 \\
$L_1$ (ep. 93)                              & 0.839 & 27.1 \\
$L_1$ (ep. 87), then VGG                    & 0.786 & 20.4  \\
$L_1$ (ep. 87), then SSIM                   & 0.845 & 26.8  \\
$L_1$ (ep. 87), then $L_1$ + 5e-5 * VGG     & 0.833 & 26.6  \\
$L_1$ (ep. 87), then $L_1$ + 1e-5 * VGG     & 0.837 & 26.8  \\
Linear interpolation                        & 0.687 & 26.7  \\
	\hline
\end{tabular}
\vspace{1em}
\caption{SSIM and PSNR metrics for different loss functions on the validation set. Note that $L_1$ (ep. 87) was trained on DAVIS16 for 37 epochs, and then for 50 epochs with DAVIS17. $L_1$ (ep. 93) is a further extension of this last model.}
\label{table:comparisonAllModels}
\end{table}

In order to compare our results with those from the original Adaptive Separable Convolution paper \cite{AdapConv}, we used the video "See you again" by Wiz Khalifa downscaled to 960x540, which is what was used in that paper for evaluation. The results of this benchmark can be seen in table \ref{table:comparisonKhalifa}. Note that these values were computed from a subset of frames of the original video (over 5k images), selecting triplets at a distance of 30 frames from each other.

We can see from table \ref{table:comparisonKhalifa} that our results are not very different from those in the original paper, despite the use of a smaller training dataset. In most cases, the quantitatively better model ($L_1$) produces better validation results in terms of SSIM and PSNR during training. However, in this particular video we produce higher scores with our combined loss in comparison with the $L_1$ loss. This tells us that our custom function produces a good trade-off between perceived visual quality and metric evaluation.  

\begin{table}[h]
\centering
\begin{tabular}{|l|c|c|}
\hline
	Model & SSIM & PSNR \\
	\hline
$L_1$, ours                         & 0.963   & 36.5  \\
Combined (qualitative), ours        & 0.966   & 37.3  \\
$L_1$, \citeauthor{AdapConv}        & 0.968   & 41.31 \\
VGG, \citeauthor{AdapConv}          & 0.965   & 40.88 \\
	\hline
\end{tabular}
\vspace{1em}
\caption{SSIM and PSNR metrics for "See you again"}
\label{table:comparisonKhalifa}
\end{table}

We also made some additional observations during visual inspection of the interpolated images, i.e. those that were generated during training as well as additional tests we ran after training.

The limitation of our kernel size is evident in multiple tests, one of which can be seen here in figure \ref{fig:scooter}. We can see that the interpolating network is not able to find an intermediate position for many of the background objects (ex. the car's front wheel) as the motion exceeds the kernel size. This also shows why the kernel does not work well on higher resolution images as the same movement occurs over a larger amount of pixels. A similar effect occurs when an object moves very close to the camera, as the magnitude of the movement increases in relation to the frame.

\begin{figure*}
    \begin{center}
    	\includegraphics[width=0.45\textwidth]{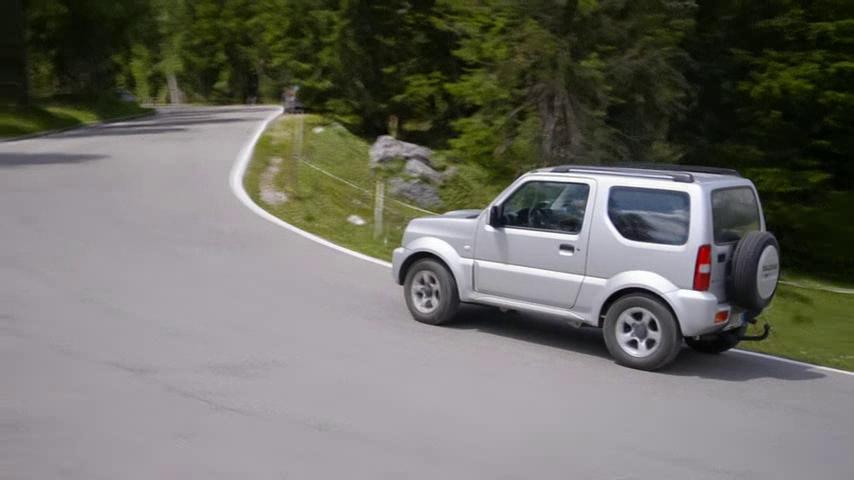}
	    \includegraphics[width=0.45\textwidth]{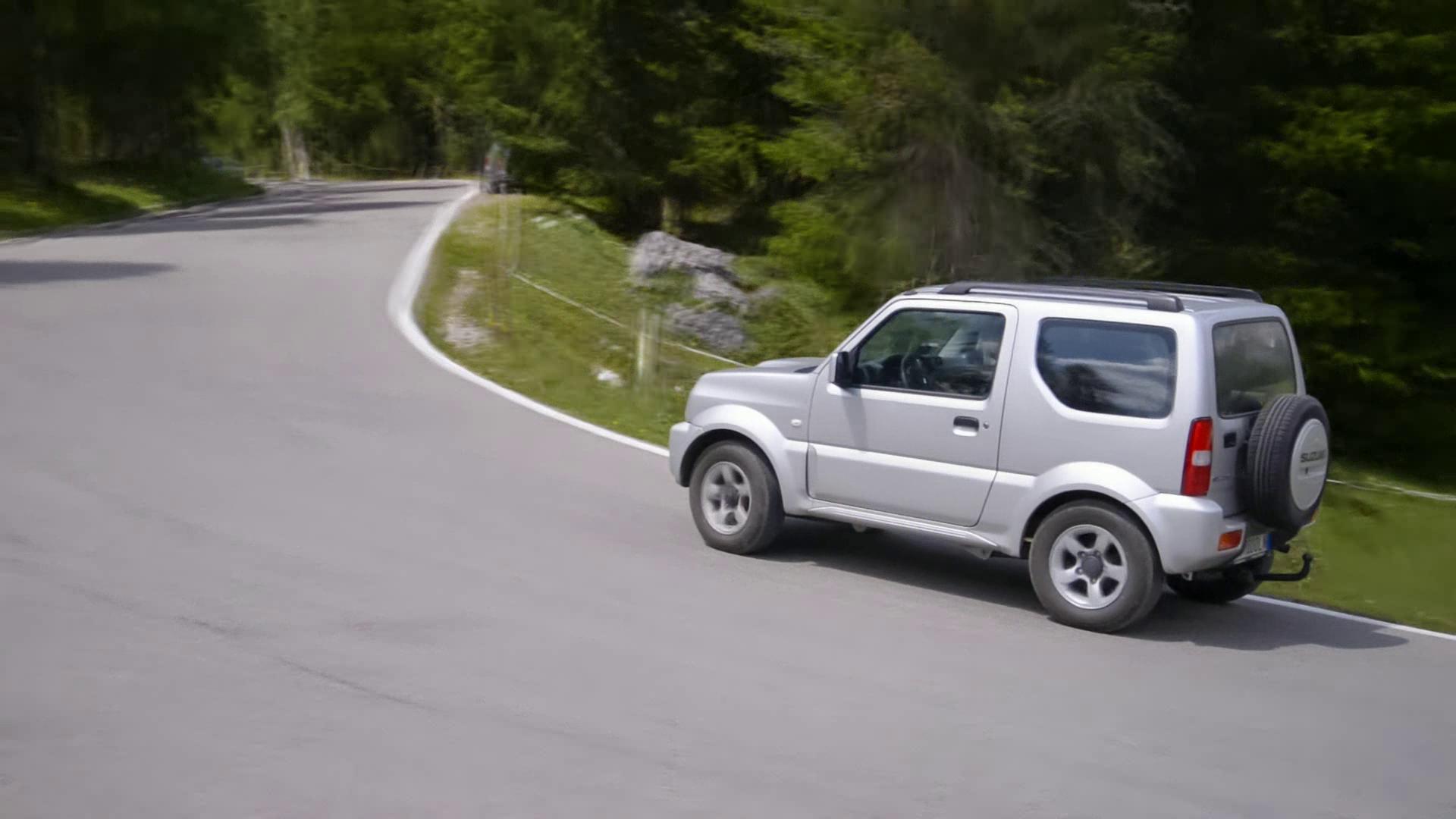}
	    \\\vspace{0.2em}
	    \adjincludegraphics[width=0.45\textwidth,trim={{0.55\width} {0.6\height} {0.2\width}  {0.15\height}},clip]{pred_car_forest_480.jpg}
	    \adjincludegraphics[width=0.45\textwidth,trim={{0.55\width} {0.6\height} {0.2\width}  {0.15\height}},clip]{pred_car_forest_1080.jpg}
    \end{center}
    \caption{Interpolated frames from 480p (left) and 1080p (right) of "Car turn" from DAVIS 17, using the combined loss model.}
    \label{fig:car_res}
\end{figure*}

An example of how resolution can have a difference in image quality is shown in figure \ref{fig:car_res}. We interpolated a frame containing a turning car with large background motion in both 480p and 1080p. As we can see, the resolution can have a strong effect on the quality of the resulting frames. The trees in the background present large distortions and lower detail in the higher-resolution version. This supports the claim made by \citeauthor{AdapConv} that the network would perform poorly in higher resolutions.

\vspace{-0.5em}
\section{Conclusion}
\label{sec:conclusion}

\vspace{-0.5em}
In this project, we have implemented all aspects of the SepConv network structure, as well as most of their data pre-processing/augmentation steps. Furthermore, we have tested loss functions that were not considered in the original work. Due to the limited resources, however, we used a dataset of reduced size in comparison to the original network by \citeauthor{SepConv}. Although our dataset was of around the same size as in other state-of-the-art approaches (ex. \cite{PhaseNet}), our experimental evidence shows that, in comparison to the results of SepConv, reducing the number of training samples had a detrimental effect on learning -- albeit not a large one. Although we have not conducted experiments to investigate the effect of the dataset size specifically, there has been a positive effect from switching from DAVIS 2016 to DAVIS 2017 after 37 epochs, most likely due to the larger size of the dataset. There are few plausible alternative explanations for the score difference, as we followed the reference training process closely. Therefore, we conclude that SepConv requires a larger dataset than other state-of-the-art methods.

Nevertheless, the results we achieved were satisfactory by both qualitative and quantitative measures. The combined loss function turned out to provide the best results by visual inspection (subjectively) and, at the same time, providing quantitative scores closer to those obtained by $L_1$. 

An inherent limitation of the SepConv network structure is the amount of motion it is able to handle. This especially impacts its ability to interpolate high-resolution videos: a 51x51 pixel area may cover a large enough space in a 480p video, but possibly not if we are dealing with 4k resolution. In fact, the authors do not recommend to use their approach on videos larger than 1280x720. This is also evident in the tests we have performed with our final models.


\vspace{0.7em}
\fakesection{Supplementary material}
\label{sec:supplementary-material}
\noindent
\large{\textbf{Supplementary material}}
\normalsize\noindent

\vspace{0.5em}
The full implementation of the network alongside additional material is available at  \url{https://github.com/martkartasev/sepconv}. 
It contains a video demonstration as well as a pre-trained model of the network.

\printbibliography


\newpage
\onecolumn
\thispagestyle{empty}
\onecolumn

\fakesection{Appendix}
\label{sec:appendix}
\noindent
\large{\textbf{Appendix}}
\normalsize\noindent
\vspace{1em}

\begin{figure}[h!]
    \begin{center}
	    \includegraphics[width=0.45\textwidth]{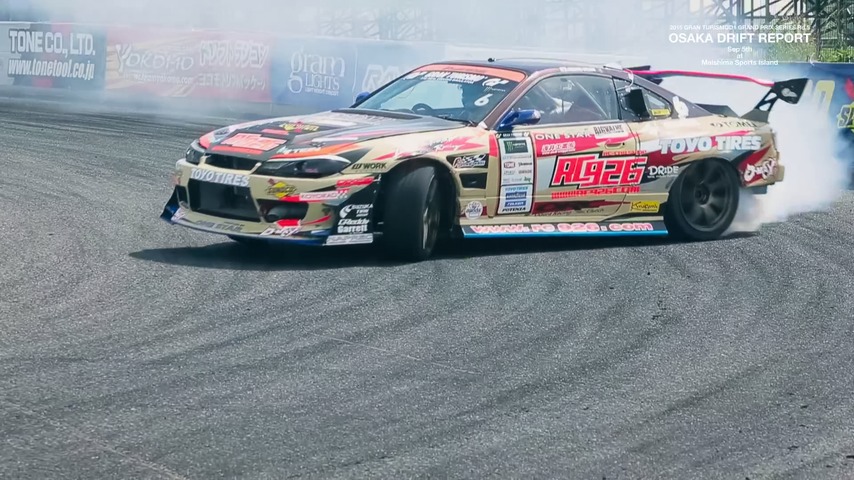}
	    \includegraphics[width=0.45\textwidth]{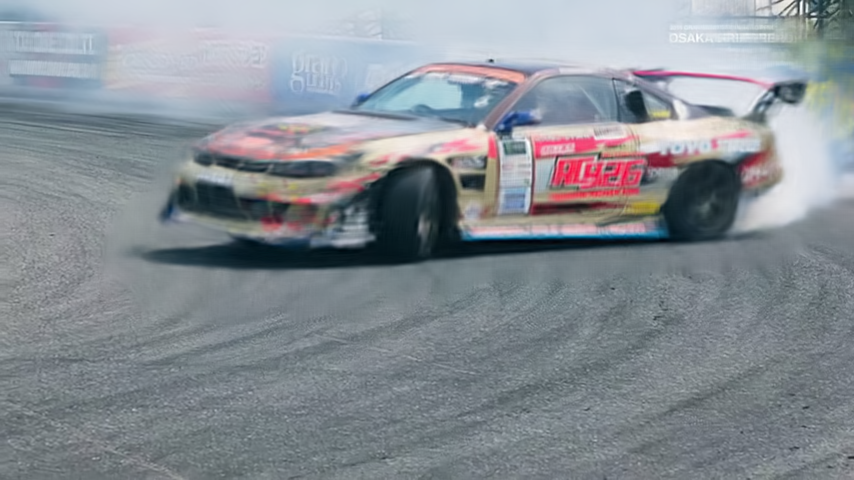}
	    \\\vspace{0.2em}
	    \adjincludegraphics[width=0.45\textwidth,trim={{0.3\width} {0.45\height} {0.4\width}  {0.25\height}},clip]{gt_car.jpg}
	    \adjincludegraphics[width=0.45\textwidth,trim={{0.3\width} {0.45\height} {0.4\width}  {0.25\height}},clip]{pred_car.png}
	    \\\vspace{0.2em}
	    \includegraphics[width=0.45\textwidth]{gt_dancer.jpg}
	    \includegraphics[width=0.45\textwidth]{pred_dancer.png}
	    \\\vspace{0.2em}
	    \adjincludegraphics[width=0.45\textwidth,trim={{0.4\width} {0.63\height} {0.3\width}  {0.07\height}},clip]{gt_dancer.jpg}
	    \adjincludegraphics[width=0.45\textwidth,trim={{0.4\width} {0.63\height} {0.3\width}  {0.07\height}},clip]{pred_dancer.png}
    \end{center}
    \caption{Ground truth vs. interpolated frame with the best qualitative model}
\end{figure}


\end{document}